\documentclass[runningheads]{llncs}
\usepackage{graphicx}
\usepackage{soul}
\usepackage{caption}
\usepackage{tikz}
\usepackage{comment}
\usepackage{amsmath,amssymb} 
\usepackage{graphicx}
\usepackage{booktabs}
\usepackage{multirow}
\usepackage{pifont}
\usepackage{bm}
\usepackage{color}
\newcommand{\cmark}{\ding{51}}%
\newcommand{\xmark}{\ding{55}}%
\newcommand{\shortmodelname}{SPOT}
\usepackage[accsupp]{axessibility}  

\usepackage{color}
\usepackage[pagebackref=true,breaklinks=true,colorlinks]{hyperref}

\newcommand{\ie}{\textit{i.e.}}
\newcommand{\eg}{\textit{e.g.}}

\begin{document}
\pagestyle{headings}
\mainmatter
\def\ECCVSubNumber{100}  

\title{Semi-Supervised Temporal Action Detection with Proposal-Free Masking} 


\titlerunning{Abbreviated paper title}
%
\titlerunning{Semi-Supervised TAD with Proposal-Free Masking}
\author{Sauradip Nag\inst{1,2} \and
Xiatian Zhu\inst{1,3} \and
Yi-Zhe Song\inst{1,2} \and
Tao Xiang\inst{1,2}}
\authorrunning{Nag et al.}
\institute{CVSSP, University of Surrey, UK \and
iFlyTek-Surrey Joint Research Centre on Artificial Intelligence, UK \and
Surrey Institute for People-Centred Artificial Intelligence, University of Surrey, UK \\
\email{\{s.nag,xiatian.zhu,y.song,t.xiang\}@surrey.ac.uk}}
\maketitle

\begin{abstract}
Existing temporal action detection (TAD) methods rely on 
a large number of training data with segment-level annotations.
Collecting and annotating such a training set is thus highly expensive and unscalable.
Semi-supervised TAD (SS-TAD) alleviates this problem by leveraging unlabeled videos freely available at scale.
However, SS-TAD is also a much more challenging problem than supervised TAD, and consequently much under-studied. Prior SS-TAD methods directly combine an existing proposal-based TAD method and a SSL method. 
Due to their {\em sequential} localization (\eg, proposal generation) and classification design, they are prone to  proposal error propagation.
To overcome this limitation, in this work we propose a novel
{\em \underline{S}emi-supervised Temporal action detection model based on
 \underline{P}rop\underline{O}sal-free \underline{T}emporal mask} (\shortmodelname) with a {\em parallel}
localization (mask generation) and classification architecture. 
Such a novel design effectively eliminates the dependence between localization and classification by cutting off the route for error propagation in-between.
We further introduce an interaction mechanism between classification and localization for prediction refinement, and a new pretext task for self-supervised model pre-training.
    Extensive experiments on two standard benchmarks show that our {\shortmodelname} outperforms state-of-the-art alternatives, often by a large margin. The PyTorch implementation of {\shortmodelname} is available at \href{https://github.com/sauradip/SPOT}{https://github.com/sauradip/SPOT}

\end{abstract}

\section{Introduction}
\label{sec:intro}
Temporal action detection (TAD) aims to predict the temporal duration (\ie, the start and end points) and the class label of each action instances in an untrimmed video \cite{idrees2017thumos,caba2015activitynet}.
Most state-of-the-art TAD methods \cite{xu2021low,xu2020g,buch2017sst,wang2017untrimmednets,zhao2017temporal,nag2022gsm,nag2021temporal} rely on training datasets containing a large number of videos (\eg, hundreds) with exhaustive segment-level annotations. Obtaining such annotations is tedious and costly. 
This has severely limited the usability of existing TAD methods in low data setting \cite{nag2021few,nag2022pclfm}.

\begin{figure}[t]
\centering
    \includegraphics[scale=0.36]{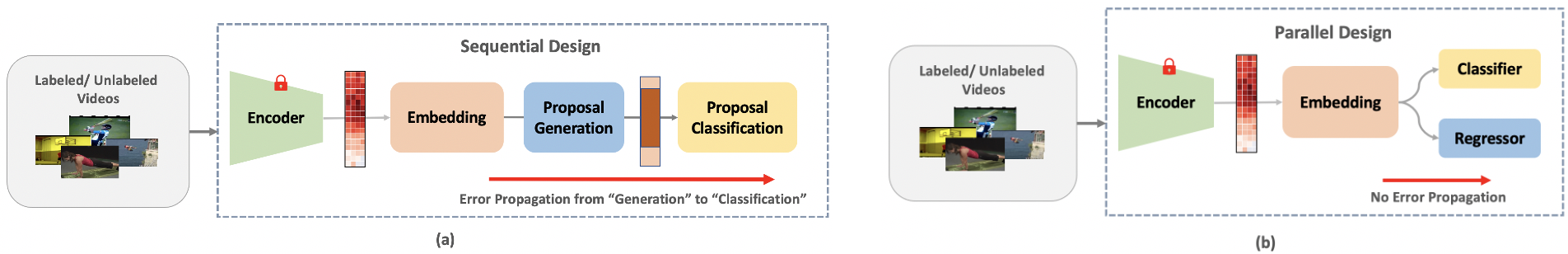}
    \caption{
    Illustration of main design difference between existing and our SS-TAD methods.
    (a) Existing SS-TAD methods suffers from the intrinsic localization error propagation problem, due to the {\em sequential} localization (\eg, proposal generation) and classification design.
    (b) We solve this problem by designing a {Proposal Free} Temporal Mask ({\shortmodelname}) learning model with a {\em parallel} localization and classification architecture.
   }
    \label{fig:intropic}
\end{figure}

Semi-supervised learning (SSL) offers a solution to the annotation cost problem by exploiting a large amount of unlabeled data along with limited labeled data \cite{tarvainen2017mean,sohn2020fixmatch}.
This has led to an emerging research interest \cite{ji2019learning,wang2021self} in  semi-supervised TAD (SS-TAD). Existing methods adopt an intuitive strategy of combining an existing TAD models, dominated by proposal-based methods and a SSL method. 
However, this strategy is intrinsically sub-optimal and prone to an error propagation problem. As illustrated in Fig.~\ref{fig:intropic}(a), this is because
existing TAD models adopt a {\em sequential} localization (\eg, proposal generation) and classification design. When extended to SSL setting, 
the localization errors, inevitable when trained with unlabeled data, can  be easily propagated to the classification module
leading to accumulated errors in class prediction.

To overcome the above limitation, in this work we propose a novel { Semi-supervised PropOsal-free Temporal Masking} ({\shortmodelname}) model with a {\em parallel}
localization (mask generation) and classification architecture (see Fig.~\ref{fig:intropic}(b)).
Concretely, {\shortmodelname} consists of a classification stream and a mask based localization stream, established in parallel on a shared feature embedding module.
This architecture design has no sequential dependence between localization
and classification as in conventional TAD models,
therefore eliminating the localization error propagation problem.
We further introduce a boundary refinement algorithm and a novel pretext task for self-supervised model pre-training. 
We integrate the {\shortmodelname} with pseudo labeling for SS-TAD
with new classification and mask loss functions formulated specifically for our parallel design. Moreover, \textcolor{black}{by virtue of being proposal free, our model is 30$\times$/2$\times$ faster in training/inference than existing alternatives.}

\noindent {\bf Contributions}. 
{\bf (1)}  To solve the localization error propagation problem suffered by existing SS-TAD methods, we propose a {\em Proposal-Free Temporal Masking} ({\shortmodelname}) model
with a new parallel classification and localization architecture\footnote{
{Note, instead of contributing a novel generic SSL algorithm, we propose a new TAD architecture designed particularly for facilitating the usage of prior SSL methods (\eg, pseudo labeling) in the sense of minimizing localization error propagation.}}.
{\bf (2)} 
We further design a novel pretext task for model pre-training 
and a boundary refinement algorithm.
{\bf (3)} Extensive experiments on two standard benchmarks (ActivityNet-V1.3 and THUMOS14) show that
the proposed {\shortmodelname} outperforms significantly alternative state-of-the-art SSL methods.
\section{Related Works}
\label{sec:related}

\noindent{\bf Temporal Action Detection }
While all existing TAD methods use action proposals, they differ in how the proposals are produced.

\noindent{\em Anchor-based proposal learning methods}
generate proposal with a pre-determined set of anchors. Inspired by object detection in static images \cite{ren2016faster},
R-C3D \cite{xu2017r} proposes to use anchor boxes.
It follows the structure of {proposal generation and classification} in design.
With a similar model design, TURN \cite{gao2017turn} aggregates local features to represent snippet-level features, which are then used for temporal boundary regression and classification.
Later, GTAN \cite{long2019gaussian}
improves the proposal feature pooling procedure with a learnable Gaussian kernel for weighted averaging.
G-TAD \cite{xu2020g}
learns semantic and temporal context via graph convolutional networks for better proposal generation. 
\textcolor{black}{Recently, VSGN \cite{zhao2021video} improves short-action localization with a cross-scale multi-level pyramidal architecture.}
Note that these anchor boxes are often exhaustively generated so are high in number. 

\noindent{\em Anchor-free proposal learning methods}
directly learn to predict temporal proposals (\ie, start and end times) 
\cite{zhao2017temporal,lin2018bsn,lin2019bmn}.
For example, SSN \cite{zhao2017temporal} decomposes an action instance into three stages (starting, course, and ending)
and employs structured temporal pyramid pooling
to generate proposals.
%
BSN \cite{lin2018bsn} predicts the start, end and actionness at each temporal location and generates proposals with high start and end probabilities.
%
Later, BMN \cite{lin2019bmn}
additionally generates a boundary-matching confidence map to improve proposal generation. 
\textcolor{black}{BSN++ \cite{su2020bsn++} further extends BMN with a complementary boundary generator to capture rich context.
CSA \cite{sridhar2021class} enriches the proposal temporal context via attention transfer.}
While no pre-defined anchor boxes are required, these methods often have to exhaustively
pair all possible locations predicted with high scores.
%
%
So both anchor-based and anchor-free TAD methods
have a large quantity of temporal proposals to evaluate. 
Critically,
both groups of TAD models in essence 
adopt a {\em sequential}
localization (mask generation) and classification architecture.
This will cause the localization error propagation problem
for SS-TAD.
Our {\shortmodelname} is designed to address this limitation by removing
the dependence between localization and classification
thus cutting off the path for error propagation.
\noindent {\bf Semi-supervised learning}
(SSL) \cite{zhu2006semi,chapelle2010semi}
%
has been widely adopted in computer vision for image classification \cite{berthelot2019mixmatch,sohn2020fixmatch,chen2018semi},
object detection \cite{Tang_2019_ICCV,Zhao_2020_CVPR}, semantic segmentation \cite{Ouali_2020_CVPR,Ibrahim_2020_CVPR}, and pose estimation \cite{Chen_2019_ICCV,Mitra_2020_CVPR}.
Two dominant learning paradigms in SSL are
%
pseudo-labeling \cite{sohn2020fixmatch,Yan_2019_ICCV,lee2013pseudo,kim2020distribution} and consistency regularization \cite{tarvainen2017mean,laine2016temporal,xie2020unsupervised,miyato2018virtual}. 
Key to pseudo-labeling is to reliably estimate the labels of unlabeled data which in turn are used to further train the model.
Instead, consistency regularization enforces the output of a model to be consistent 
at the presence of variations in the input space and/or the model space. The variations can be implemented by adding noises, perturbations or forming multiple variants of the same data sample or the model. 
In this work, 
we focus on designing a TAD model particularly suitable for SSL, while following the pseudo-labelling paradigm for exploiting unlabeled data for training.
%

\noindent {\bf Semi-supervised Temporal Action Detection} (SS-TAD).
SSL has only been studied very recently in the context of TAD. Existing SS-TAD works \cite{ji2019learning,wang2021self,shi2021temporal} naively combine existing semi-supervised learning
and TAD  methods.
They are thus particularly prone to the aforementioned localization error propagation problem when trained with unlabeled data.
We solve this problem for the first time
by introducing a novel proposal-free temporal mask learning model.
%

\noindent{\bf Self-supervised learning}
aims to learn generic feature representations 
from a large amount of unlabeled data \cite{chen2020simple,he2019momentum,grill2020bootstrap}.
It is typically designed to provide a pre-trained model
for a downstream task to further fine-tune
with a specific labeled training data.
We have seen a recent surge of self-supervised learning studies with a focus on both object recognition in images \cite{wu2018unsupervised,chen2020simple,he2019momentum,grill2020bootstrap,misra2020self,vondrick2018tracking,chen2020exploring} and action classification in videos
\cite{alwassel2020xdc,benaim2020speednet,miech20endtoend,shuffle_learn_eccv16,arrow_cvpr18}.
The most related works to our self-supervised learning based pre-training are
very recently introduced in \cite{xu2021low,xu2021boundary}. They aim to improve the video encoder for the fully supervised TAD problem.
In contrast, we focus on pre-training a superior TAD head 
in the context of semi-supervised learning.

\section{Proposal-Free Temporal Mask Learning}
\label{sec:method}
\begin{figure*}[h]
\begin{center}
  \includegraphics[scale=0.37]{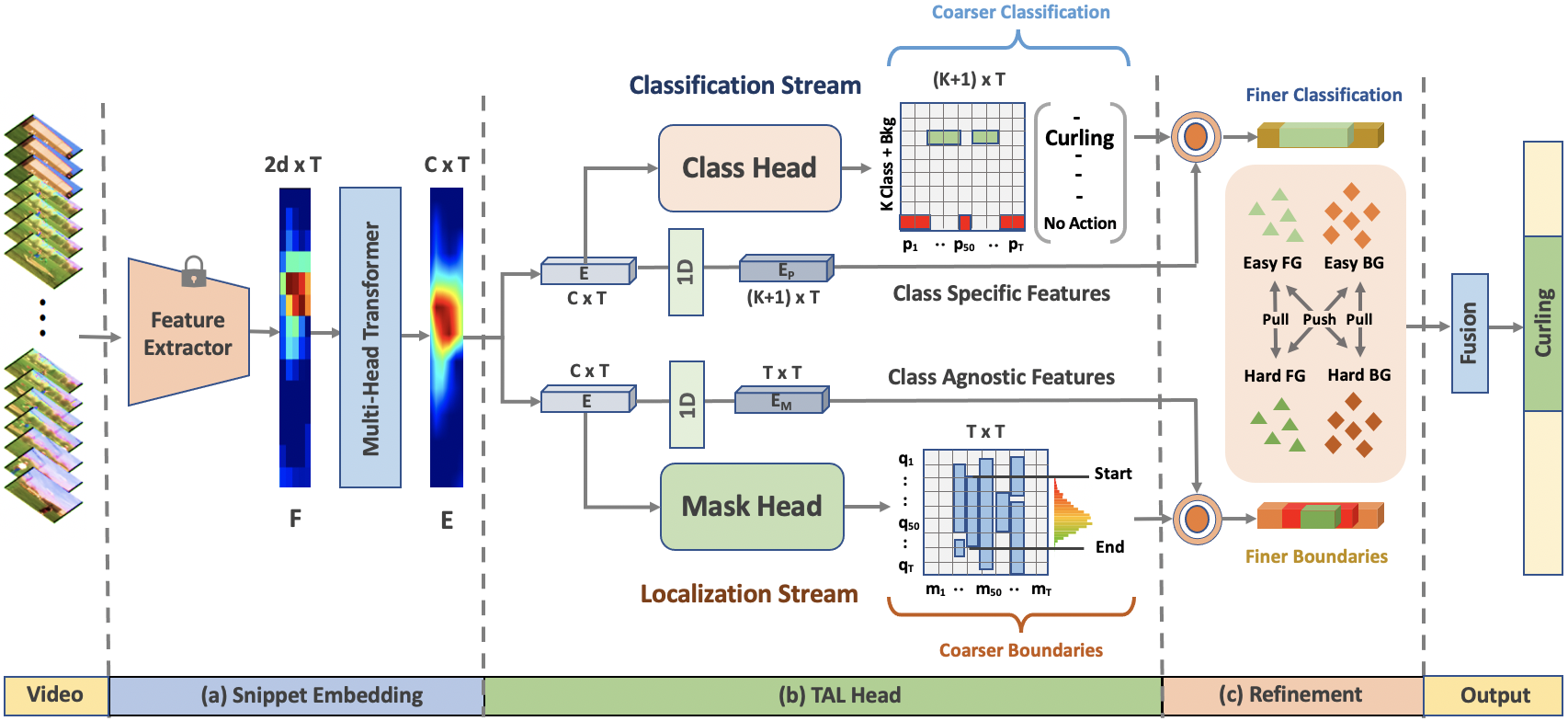}
\end{center}
\caption{\textbf{Overview of the proposed {\em \underline{S}emi-supervised Temporal action detection model based on
 \underline{P}rop\underline{O}sal-free \underline{T}emporal mask} (\shortmodelname).} Given an untrimmed video $V$, (a) we first extract a sequence of $T$ snippet features with a pre-trained video encoder and conduct self-attention learning to obtain the snippet embedding $E$ with a global context.
(b) For each snippet embedding, we then predict a classification score $P$ with the classification stream and a foreground mask $M$ with the mask stream in parallel,
(c) both of which are further used for boundary refinement.
It is based on mining hard\&easy foreground (FG) and background (BG) snippets.
For SS-TAD, we alternatingly predict and leverage pseudo class and mask labels of unlabeled training videos,
along with labeled videos.
}
\label{fig:network}
\end{figure*}

\noindent{\bf Approach overview }
In semi-supervised temporal action detection (SS-TAD), 
we have access to a small set of $N_{l}$ labeled videos 
$\mathcal{D}^{l}= \{V_i, \Psi_i\}_{i=1}^{N_{l}}$
and a large set of $N_{u}$ unlabeled videos $\mathcal{D}^{u} = \{U_i\}_{i=1}^{N_{u}}$.
Each labeled video's annotation $\Psi_i = \{(\psi_j, \xi_j, y_j)\}_{j=1}^{M_i}$ 
contains the start time $\psi_{j}$, the end time $\xi_{j}$, and  the class label $y_j \in \mathcal{Y}$ for each of $M_i$ action instances. 
We denote the label space as $\mathcal{Y}=[1,\cdots,K+1]$ with $K$ action and one background classes.
For more effective SS-TAD, we propose a {\em Proposal-Free Temporal Mask} ({\shortmodelname}) learning method (see Fig.~\ref{fig:network}).
It has two components:
video snippet embedding (Sec.~\ref{sec:emb}),
and TAD head (Sec.~\ref{sec:tal}).
The latter is our core contribution.

\subsection{Video Snippet Embedding}
\label{sec:emb}
\textcolor{black}{
Given a varying length untrimmed video $V$,
following the standard practice \cite{xu2020g,lin2019bmn} we first sample $T$ equidistantly distributed temporal snippets (points) over the entire length and \textcolor{black}{use a  fine-tuned two-stream video encoder similar to
\cite{lin2019bmn} to extract 
RGB $X_{r} \in \mathbb{R}^{d \times T}$ and optical flow features $X_{o} \in \mathbb{R}^{d \times T}$ at the snippet level, where $d$ denotes the feature dimension. 
We then concatenate them as $F=[ X_{r};X_{o}] \in \mathbb{R}^{2d \times T}$.
Each snippet is a short sequence of (16 in this work) consecutive frames. }}
While $F$ contains local spatio-temporal information, 
it lacks a global context critical for TAD.
We hence leverage the self-attention mechanism \cite{vaswani2017attention} to learn the global context.
Formally, we set the input $\{query, key, value\}$ of a multi-head Transformer encoder
$\mathcal{T}()$ as the features $\{F, F, F\}$ (Fig.~\ref{fig:network}(a)). \textcolor{black}{Positional encoding is not applied as it is found to be detrimental (see Appendix C in Supplementary)}
The final snippet embedding is then obtained as
$E = \mathcal{T}(F) \in \mathbb{R}^{C \times T}$ with $C$ being the embedding dimension.

\subsection{TAD Head}
\label{sec:tal}
To realize a proposal-free design,
we introduce a temporal mask learning based TAD head.
It consists of two parallel streams (Fig.~\ref{fig:network}(b)):
one for snippet classification 
and 
the other for temporal mask inference.
This design breaks the sequential dependence
between localization and classification
which causes unwanted error propagation in the existing TAD models.

\noindent \textbf{Snippet classification stream }
Given the $t$-th snippet $E(t) \in \mathbb{R}^{c}$ (\ie, the $t$-th column of $E$),
our classification branch predicts a probability distribution $\bm{p}_t \in \mathbb{R}^{(K+1)\times 1}$ over $\mathcal{Y}$.
This is realized by a 1-D convolution layer $H_c$ followed by 
a softmax normalization.
For a video with $T$ snippets, the output of the classification branch can be expressed as:
\begin{equation}\label{eqn_3}
\bm{P} : = softmax\Big(H_c(E)\Big) \in \mathbb{R}^{(K+1) \times T}.
\end{equation}

\noindent \textbf{Temporal mask stream } 
\textcolor{black}{In parallel to the classification stream, this stream predicts temporal masks of action instances across the whole temporal span of the video.}
Given the $t$-th snippet $E(t)$, 
it outputs a mask vector $\bm{m}_t = [q_{1}, \cdots, q_{T}] \in \mathbb{R}^{T \times 1}$
with each element $q_{i} \in [0, 1]$ ($i \in [1, T]$) indicating the foreground probability of the $i$-th snippet \textcolor{black}{(see Fig. 2(b) for illustration)}.
%
This is implemented
by a stack of three 1-D convolution layers $H_m$ as:
\begin{equation}\label{eqn_4}
\bm{M} : = sigmoid\Big(H_m(E)\Big) \in \mathbb{R}^{T \times T}, 
\end{equation}
where the $t$-th column of $\bm{M}$ is the temporal mask prediction by the $t$-th snippet.
With the proposed mask signal as the model output supervision,
proposals are no longer required for facilitating 
SS-TAD learning.


\begin{figure}[h]
\centering
    \includegraphics[scale=0.32]{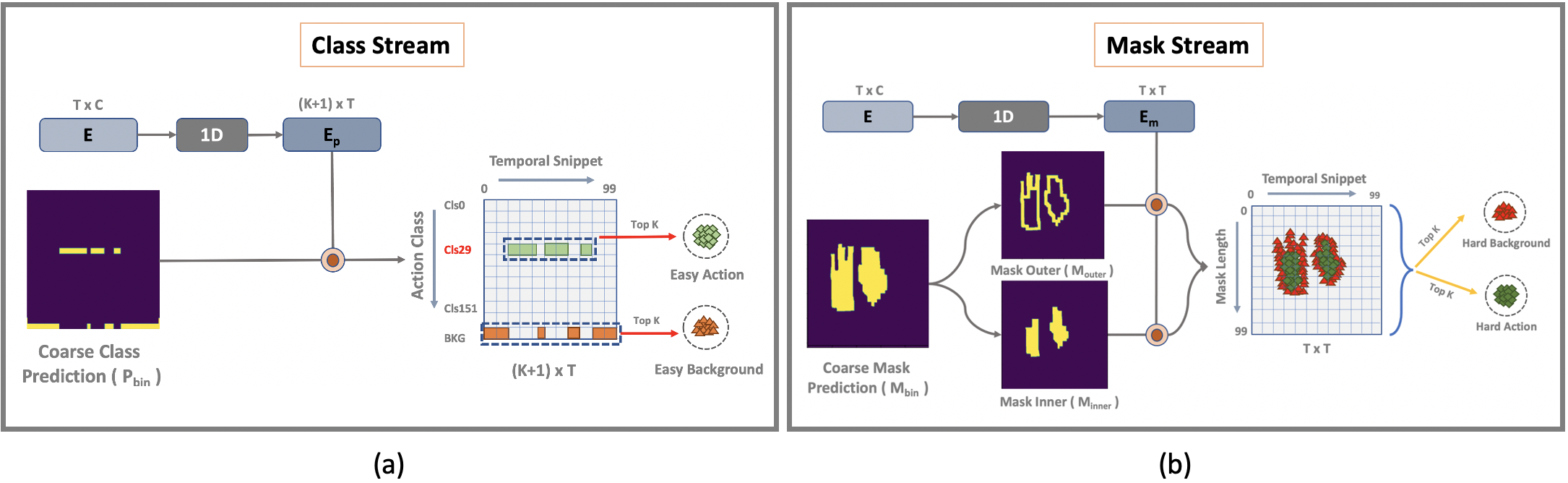}
    \vskip -0.3cm
    \caption{Snippet selection for inter-stream interaction.}
    \label{fig:consnip}
\end{figure}

\noindent{\bf Boundary Refinement}
TAD methods typically struggle at estimating accurately the boundary between foreground and background segments. This problem is more pronounced for a SS-TAD model by learning from unlabeled data.
To mitigate this problem,
we design an inter-stream interaction mechanism
for TAD refinement at the end of the TAD head (Fig.~\ref{fig:network}(c)).
\textcolor{black}{
More specifically,
we focus on ambiguous snippets in the transition between foreground and background (\ie, temporal boundary). 
They are considered as {\em hard} snippets,
in contrast to those easy ones
located inside mask or
background intervals far away from temporal borders
with more trustworthy interference. %
We detect hard snippets by examining the structure of temporal mask $\bm{M}$.}
First, we threshold the $M$ to obtain a binary mask 
as $M_{bin} := \eta(M - \theta_{m})$,
where $\eta(.)$ is the Heavyside step function and $\theta_{m}$ is the threshold.
\textcolor{black}{As shown in Fig.~\ref{fig:consnip}(b), we consider the snippets spanned by the eroded mask boundary as {\em hard background} whereas 
those by the non-boundary mask as {\em hard foreground}. We use a differentiable morphological erosion \cite{riba2020kornia} to obtain the eroded mask $M_{outer}$.
We denote the complement of $M_{outer}$ as $M_{inner}$.
They are formally defined as: 
\begin{equation*}
     M_{outer} = \mathbb{E}(M_{bin}, e), \;\;\;\;
    M_{inner} = M_{bin} - M_{outer},
\end{equation*}
where $\mathbb{E(.)}$ is the differentiable erosion operation and $e$ is the erosion kernel size. We represent the top k scoring hard snippets by multiplying the downsampled embedding $E_{m}$ (obtained after applying 1-D convolution on the embedding $E$).
They are obtained as:
\begin{equation*}
    X^{fg} = topk(M_{inner}*E_{m}), \;\;\;\;
    X^{bg} = topk(M_{outer}*E_{m}), 
\end{equation*}}
\noindent\textcolor{black}{{As} the snippets from the same instance often predict different length 1-D masks, applying a binarization process on top generates masks of arbitrary shapes, as illustrated in Fig.~3(a). We calculate the boundary features for all of these mask clusters and select the top snippets from all of the clusters jointly.}
For inter-stream interaction, 
\textcolor{black}{
we further use high-confidence foreground and background snippets from the classification stream. 
We thus consider them as \textit{easy foreground} and \textit{easy background} due to their {\em easy to predict} property. }
As seen in Fig.~\ref{fig:consnip}(a),
we similarly select top scoring foreground and background snippets from the thresholded classification output $P_{bin} := \eta(P - \theta_{c})$ as:
\begin{align*}
    Y^{fg} = topk(argmax((P_{bin}*E_{p})[:K,:])), \;\;
    Y^{bg} = topk((P_{bin}*E_{p})[K+1,:]). 
\end{align*}
where $E_{p}$ is obtained by passing the embedding $E$ into a 1-D conv layer for matching the dimension of $P$.
We adopt infoNCE loss \cite{he2020momentum} for TAD refinement:
\begin{equation}
    L_{ref} = tri(x^{fg}, y^{fg}, y^{bg}) + 
    tri(y^{bg}, x^{bg}, y^{fg}),
    \label{eq:ref}
\end{equation}
where $x^{*} \in X^*$ and $y^* \in Y^*$ and $tri(\cdot, \cdot, \cdot)$
defines the foreground and background triplet training samples.
\textcolor{black}{In this way, we maximize the correlation between easy and hard snippets of the same category (foreground or background),
refining the inter-stream feature representation for
better perceiving temporal boundaries.}

\begin{figure}[t]
\centering
    \includegraphics[scale=0.16]{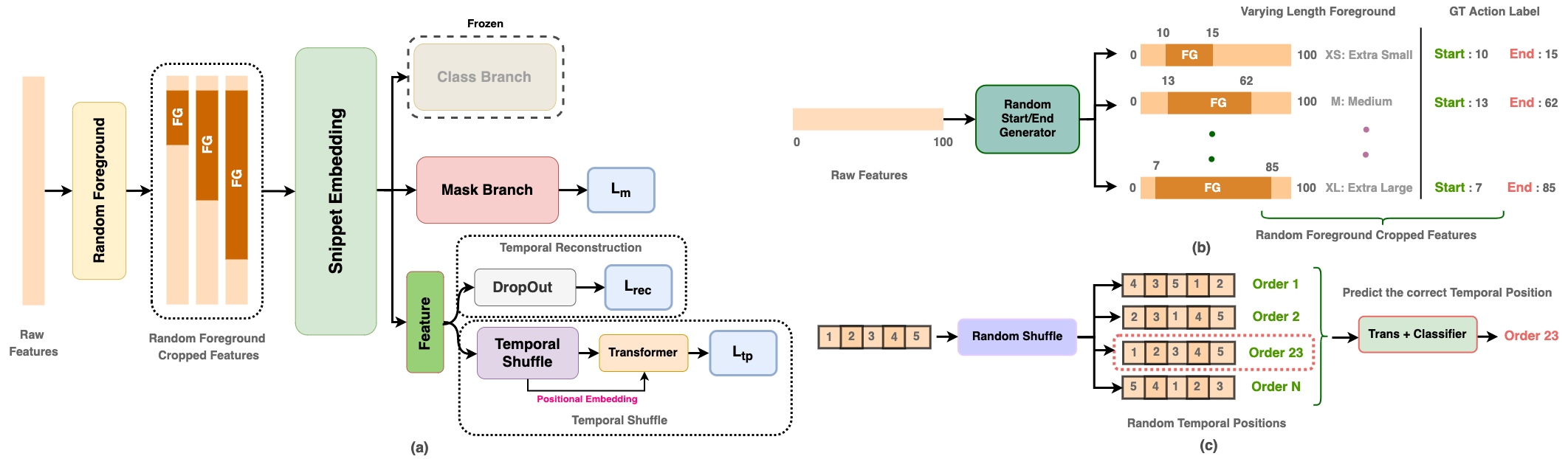}
    \caption{Illustration of our self-supervised pre-training.}
    \label{fig:pretext}
\end{figure}

\subsection{Model Training}
To better exploit unlabeled data,
we formulate a two-staged training pipeline,
including self-supervised pre-training and semi-supervised fine-tuning.

\subsubsection{Stage I: Self-supervised pre-training}
We introduce a pretext task based on a novel notion of {\em random foreground}
specially designed for TAD model pre-training.
Given a video feature sequence $F \in \mathbb{R}^{T\times d}$,
we sample a random \textcolor{black}{token} segment $(s, e)$ of varying proportions as foreground and the remaining \textcolor{black}{tokens} as background
(Fig.~\ref{fig:pretext}).
\textcolor{black}{Cropping out foreground at feature level has shown to be useful in learning discriminative representation \cite{patrick2021space}. Motivated by this, we zero out the background snippet features whilst keeping for the pseudo foreground.}
\textcolor{black}{This shares a similar spirit with masked region modeling \cite{chen2020uniter,bao2021beit}
with a different objective of detecting the location of masked segment.}
With such a masked feature sequence, our pretext task aims to predict jointly (1) the temporal mask with the start $s$ and end $e$ (Fig.~\ref{fig:pretext}(b)),
(2) the temporal position of each snippet after temporal shuffling (Fig.~\ref{fig:pretext}(c)),
and 
(3) the reconstruction of snippet feature (Fig.~\ref{fig:pretext}(a)).
We treat each of $T$ temporal positions as
a specific \textcolor{black}{positional class}, and apply a small Transformer with learnable positional embedding on the shuffled snippet sequences.
{\color{black} All zeroed snippet features will become {\em non-zero} after the transformer's processing, preventing the model from learning a trivial solution.}
We use the cross-entropy loss $L_{tp}$ for temporal position prediction.
\textcolor{black}{This supervision aims to learn the intrinsic temporal structure information in videos.}
The feature reconstruction loss $L_{rec}$ is measured by  $L_2$ normalized mean squared error between the backbone feature $F$ and the embedding $E$ \textcolor{black}{from the transformer. The motivation is to preserve the video encoder's discriminative information in learning global context.
By dropping random snippets, the Transformer is forced to  aggregate and utilize information from the context to predict the dropped snippets. As a result, the model can learn temporal semantic relations and discriminative features useful for TAD.}
The pretext task loss for pre-training is formulated as:
\begin{equation}
    L_{pre} = L_{m} + \lambda_{1} L_{rec} + \lambda_{2} L_{tp},
    \label{eq:pre_loss}
\end{equation}
where $L_{m}$ is the mask learning loss as defined in Eq. \eqref{eq:mask_loss} and $\lambda_{1}$ and $\lambda_{2}$ are two hyper-parameters set as 0.8 and 0.4 respectively. 
We use both labeled and unlabeled data for {\shortmodelname} pre-training
without using any ground-truth labels.

\subsubsection{Stage II: Semi-supervised fine-tuning}
\label{sec:train_s2}
We implement temporal mask semi-supervised learning
following the pseudo label paradigm \cite{sohn2020fixmatch}.
Concretely, we alternate between predicting and applying pseudo labels,
starting by using the labeled samples alone.

\noindent {\bf Pseudo label prediction}
Given an unlabeled video embedding $E$, the pseudo class label is obtained by:  
\begin{align}
    \hat{y} = max\Big( softmax(H_{c}(E)/\tau_{c}) \Big),
\end{align}
where the sharpening operator $\tau_{c} = \tau - (\tau -1)\hat{y}_{max}^{'}$ with $\hat{y}_{max}^{'}$ the maximum probability over the $K$ classes, 
and $\tau$ a hyper-parameter controlling the largest sharpening strength. 
Similarly, we obtain the pseudo mask label $\hat{g}$ as:
\begin{align}
    \hat{g} = sig\Big(H_{m}(E)/\tau_{m}\Big),
\end{align}
where $\tau_{m}$ is the mask sharpening operator,
and $sig()$ is the sigmoid function \cite{little1974existence}.
Then, we threshold the pseudo class and mask labels
at $\theta_{c}/\theta_{m}$ to be binary.

\noindent {\bf Loss functions }
For SS-TAD, we use both pseudo and ground-truth labels to 
minimize the objective loss function as formulated below.
For the {\bf \em classification} stream, we devise a class-balanced loss to tackle the intrinsic class imbalance challenge with TAD.
Inspired by \cite{wang2021adaptive,dong2019single},
we adopt the sigmoid regression based binary cross-entropy loss function.
Given a snippet $E(t)$ with class label $y$,
the classification loss function is designed as:
\begin{equation}
{L}_{bce} = 
-\log(p_{y}) - \sum_{k\neq y, k\in \mathcal{Y}}
\Big( \log(1-p_k)\Big),
\label{eq:LR}
\end{equation}
where $p_{y}$ denotes the the logits from class stream $P$.
Critically, this binary cross-entropy loss gives us the flexibility to regulate each class individually per training sample. 
This is useful because untrimmed videos often encompass a dominant proportion of background content, which would overwhelm the under-represented tail action classes $\mathcal{Y}_t$ with small training data.
To mitigate this problem, we further improve this loss by encouraging the activation of tail action classes.
Concretely, given a {\em background} snippet,
we still allow tail action classes to be activated under a certain degree $\varepsilon $ $($set as $\theta_{c})$.
This is realized by introducing a weighting mechanism as:
\begin{align}
{L}_{wbce}  = 
-\log(p_y) - \sum_{k\neq y, k\in \mathcal{Y}}
\Big( w_k \log(1-p_k)\Big) \\
\text{where} \;\; \omega_{k} = \left\{\begin{matrix}
 &  0 & if \; k \in \mathcal{Y}_t \; \text{and} \; p(k) < \varepsilon \\ 
 & 1 & otherwise
\end{matrix}\right.
\end{align}
Given a video with the background $T_{bg}$ and foreground $T_{fg}$ snippets,
our final classification loss is expressed as:
\begin{equation}\label{eq:final_clf}
    L_c = \frac{1}{T} \Big( \sum_{t \in T_{fg}}
    L_{bce}(t) + \sum_{t \in T_{bg}}L_{wbce}(t) \Big).
\end{equation}

For the {\bf \em mask} stream, we exploit a weighted binary cross-entropy loss $L_{mce}$ 
for balancing foreground and background classes, along with a binary dice loss $L_{dice}$ \cite{milletari2016v}.
For a snippet location $t$, 
we denote $\bm{m}(t)\in \mathbb{R}^{T\times 1}$ and $\bm{g}(t) \in \mathbb{R}^{T\times 1}$ as the predicted and ground-truth mask.
The mask learning loss is designed as:

\begin{equation}
\centering \footnotesize
\begin{aligned}
    & L_m = \beta_{fg}
    \sum_{t=1}^{T}
    \bm{g}(t) \log(\bm{m}(t)) + 
    \beta_{bg}
    \sum_{t=1}^{T} (1-\bm{g}(t)) \log(1-\bm{m}(t)) 
    \\
   &   \hspace{0.3in}  + \lambda_{d}
    \Big( 1 - 
    \frac{\bm{m}^\top \bm{g}}
    {\sum_{t=1}^{T} \big(\bm{m}(t)^2 + \bm{g}(t)^2 \big)} 
    \Big),
\end{aligned}
\label{eq:mask_loss}
\end{equation}
where $\beta_{fg}$ and $\beta_{bg}$
are the inverse proportion of foreground and background
used for class balance, and
$\lambda_d$ is the dice loss weight which is empirically set as 0.6.

\noindent{\em Overall learning objective}
The overall objective loss is designed as $L = L_{c} + L_{m} + L_{ref} + L_{rec}$
where $L_{ref}$ is the refinement loss (Eq.~\eqref{eq:ref}) and $L_{rec}$ is the feature reconstruction loss as described in the pre-training stage.
This loss is applied on both ground-truth and pseudo labels
for fine-tuning {\shortmodelname}. \textcolor{black}{Note that the temporal ordering loss term $L_{tp}$ is not used during fine-tuning as it gives  performance drop (see Appendix C in Supplementary). A plausible cause is its incompatibility with the fine-tuning loss $L_{c}$.}

\subsection{Model Inference}
\textcolor{black}{
At test time, we generate action instance predictions for each test video by the classification $\bm{P}$ and mask $\bm{M}$ predictions.
For $\bm{P}$, we only consider the snippets whose class probabilities are greater than $\theta_{c}$ and select top scoring snippets.
For each such top scoring action snippet,
we then obtain the
temporal mask by thresholding the $t_i$-th column of $\bm{M}$ using the localization threshold ${\Theta}$. To produce sufficient candidates,
we use a set of thresholds $\Theta=\{\theta_i\}$. 
For each candidate, we compute a confidence score $s$
by multiplying the classification and max-mask scores.
SoftNMS \cite{bodla2017soft} is finally applied to obtain top scoring results.
}

\section{Experiments}
\label{sec:exp}
\noindent \textbf{Datasets } We use two standard TAD datasets in our evaluation.
\textbf{(1)} {\em ActivityNet v1.3} is a large-scale benchmark containing 19,994 untrimmed videos with 200 classes. We adopt the standard 2:1:1 training/validation/testing video data split. 
%
\textbf{(2)} {\em THUMOS14} provides 
200 validation videos and 213 testing videos
labeled with temporal annotations for action understanding. 
We train our model on the validation set and evaluate on the test set.

\noindent{\bf Implementation details } 
For fair comparisons, \textcolor{black}{we use both TSN \cite{wang2016temporal} and I3D \cite{carreira2017quo} features in our evaluations. For ActivityNet, we use fine-tuned TSN features for fair comparison with \cite{wang2021self,lin2019bmn}. For THUMOS, we use TSN and I3D features both pre-trained on Kinetics \cite{xiong2016cuhk}.}
%
\textcolor{black}{The temporal dimension $T$ is fixed at 100/256 for ActivityNet/THUMOS respectively (see the suppl. for more analysis).}
\textcolor{black}{For AcitivityNet/THUMOS,
we first pre-train {\shortmodelname} (except the classification stream) on the training set including the unlabeled samples for 12 epochs and
then we fine-tune {\shortmodelname} for 15 epochs with a learning rate of $10^{-4}/10^{-5}$, a weight decay of $10^{-3}/10^{-5}$.}
 For boundary refinement, we set \textcolor{black}{top-$k=40$ snippets},
 $\theta_{c}/\theta_{m}$ is set as $0.3/0.7$ and $e$ is set as 7. In semi-supervised setting, the label sharpening operator $\tau$ is set as 1.1 and $\tau_{m}$ is set as 0.7. In testing, we set the threshold set for mask
$\theta = \{0.1 \sim 0.9\}$  with a step 0.1. The SoftNMS is performed on ActivityNet/Thumos with a threshold of $0.6/0.4$.

%

\begin{table*}[h]
\centering
\begin{tabular}{@{}c|c|cccc|cccccc@{}}
\toprule
\multirow{2}{*}{Labels} & \multirow{2}{*}{Methods} & \multicolumn{4}{c|}{ActivityNet}                                                  & \multicolumn{6}{c}{THUMOS}                                                                                         \\ \cmidrule(l){3-12} 
                        &                          & 0.5           & 0.75          & \multicolumn{1}{c|}{0.95}         & Avg           & 0.3           & 0.4           & 0.5           & 0.6           & \multicolumn{1}{c|}{0.7}           & Avg           \\ \midrule
\multirow{6}{*}{60\%}   & BMN$^*$ \cite{lin2019bmn}        & 47.6          & 31.7          & \multicolumn{1}{c|}{7.5}          & 31.5          & 50.8          & 45.9          & 34.8          & 23.7          & \multicolumn{1}{c|}{16.3}          & 34.3          \\ \cmidrule(l){2-12} 
                        & Mean Teacher \cite{tarvainen2017mean}+BMN            
                        & 48.0          & 32.1          & \multicolumn{1}{c|}{7.4}      
                        & 31.9          & 53.5          & 45.0          & 36.9          & 27.4          & \multicolumn{1}{c|}{19.0}          & 35.8          \\
                        & FixMatch  \cite{sohn2020fixmatch}+BMN              
                        & 48.7          & 32.9          & \multicolumn{1}{c|}{7.7} & 32.8          & 53.8          & 46.2          & 37.8          & 28.7          & \multicolumn{1}{c|}{19.5}          & 36.9          \\
                        & SSP \cite{ji2019learning}    & 49.8          & 34.5          & \multicolumn{1}{c|}{7.0}          & 33.5          & 53.2          & 46.8          & 39.3          & 29.7          & \multicolumn{1}{c|}{19.8}          & 37.8          \\
                        & SSTAP \cite{wang2021self}        & 50.1          & 34.9          & \multicolumn{1}{c|}{7.4}          & 34.0          & 56.4          & 49.5          & 41.0          & 30.9          & \multicolumn{1}{c|}{21.6}          & 39.9          \\ \cmidrule(l){2-12} 
                        & \textbf{{\shortmodelname} (Ours)}            & \textbf{52.8} & \textbf{35.0} & \multicolumn{1}{c|}{\bf 8.1}          & \textbf{35.2} & \textbf{58.9} & \textbf{50.1} & \textbf{42.3} & \textbf{33.5} & \multicolumn{1}{c|}{\textbf{22.9}} & \textbf{41.5} \\ \midrule
\multirow{6}{*}{10\%}   & BMN$^*$ \cite{lin2019bmn}        & 35.4          & 26.4          & \multicolumn{1}{c|}{8.0}          & 25.8          & 38.3          & 28.3          & 18.8          & 11.4          & \multicolumn{1}{c|}{5.6}           & 20.5          \\ \cmidrule(l){2-12} 
                        & Mean Teacher \cite{tarvainen2017mean}+BMN             & 36.0          & 27.2          & \multicolumn{1}{c|}{7.4}          & 26.6          &  41.2           & 32.1           & 23.1           & 15.0           & \multicolumn{1}{c|}{7.0}          & 23.7          \\
                        & FixMatch  \cite{sohn2020fixmatch}+BMN                & 36.8          & 27.9          & \multicolumn{1}{c|}{8.0} & 26.9          & 42.0           & 32.8           & 23.0           & 15.9          & \multicolumn{1}{c|}{8.5}          & 24.3          \\
                        & SSP \cite{ji2019learning}     & 38.9          & 28.7          & \multicolumn{1}{c|}{8.4}          & 27.6          & 44.2          & 34.1          & 24.6          & 16.9          & \multicolumn{1}{c|}{9.3}           & 25.8          \\
                        & SSTAP \cite{wang2021self}              & 40.7          & 29.6          & \multicolumn{1}{c|}{\bf 9.0}          & 28.2          & 45.6          & 35.2          & 26.3          & 17.5          & \multicolumn{1}{c|}{10.7}          & 27.0          \\ \cmidrule(l){2-12} 
                        & \textbf{{\shortmodelname} (Ours)}            & \textbf{49.9} & \textbf{31.1} & \multicolumn{1}{c|}{ 8.3}          & \textbf{32.1} & \textbf{49.4} & \textbf{40.4} & \textbf{31.5} & \textbf{22.9} & \multicolumn{1}{c|}{\textbf{12.4}} & \textbf{31.3} \\ \bottomrule
\end{tabular}
\caption{{\bf SS-TAD results on the validation set of
ActivityNet v1.3 and the test set of THUMOS14.}
Note: {\bf \em All methods except {\shortmodelname}} use the UNet \cite{wang2017untrimmednets} trained with {\bf \em 100\% class labels} for proposal classification;
These methods hence benefit from {\bf \em extra classification supervision} in comparison to {\shortmodelname}.
$^*$: Using only labeled training set.}
\vspace{-0.3in}
\label{Tab:semi}
\end{table*}

\subsection{Comparative Results}

\noindent{\bf Setting }
We introduce two SS-TAD settings with different label sizes.
For each dataset, we randomly select 10\% or 60\% training videos
as the labeled set and the remaining as the unlabeled set.
Both labeled and unlabeled sets are accessible for SS-TAD model training.

\noindent{\bf Competitors }
We compared with the following methods.
{\bf (1)} \textcolor{black}{Two} state-of-the-art supervised TAD methods:
BMN \cite{lin2019bmn}+UNet \cite{wang2017untrimmednets} and \textcolor{black}{GTAD \cite{xu2020boundary}+UNet \cite{wang2017untrimmednets}}
{\bf (2)} Two SSL+TAD methods:
As SS-TAD is a new problem, we need to implement the competitors by 
extending existing SSL methods to TAD by ourselves.
We select two top SSL methods (Mean Teacher \cite{tarvainen2017mean} and FixMatch \cite{sohn2020fixmatch}), and a state-of-the-art TAD model based on a popular 
proposal generation method BMN \cite{lin2019bmn} (\textcolor{black}{using TSN features}) and \textcolor{black}{GTAD \cite{xu2020boundary} (using I3D features)}. \textcolor{black}{Both of these models use a common untrimmed video classification model UNet} \cite{wang2017untrimmednets}.
For FixMatch \cite{sohn2020fixmatch}, \textcolor{black}{we use temporal flip (\ie, playing the video backwards) as strong augmentation and temporal feature shift as weak augmentation.}
For UNet, due to the lack of Caffe based training environment, we can only apply the official weights trained with 100\% supervision\footnote{We use official UNet weights at \url{https://github.com/wanglimin/UntrimmedNet}.}.
%
{\bf (3)} Two recent semi-supervised temporal proposal generation methods \cite{ji2019learning,wang2021self}. Note that the concurrent SS-TAD work \cite{shi2021temporal} does not give reproducible experiment settings
and prevents an exact comparison. 
Besides, \cite{shi2021temporal} achieves 
only 12.27\% avg mAP on ActivityNet v1.2 
{\em vs.} 32.1\% by {\shortmodelname} on v1.3 (v1.2 has only half classes of v1.3 hence simpler),
thus significantly inferior to the two compared methods \cite{ji2019learning,wang2021self}.
\noindent{\bf Results }
The SS-TAD results are reported in Table \ref{Tab:semi} \textcolor{black}{and Table \ref{tab:SSL_2}.}
We make the following observations:
{\bf(1)} Without leveraging unlabeled video data,
the state-of-the-art fully supervised TAD method BMN \cite{lin2019bmn} (with UNet \cite{wang2017untrimmednets}) achieves the poorest result among all the methods.
This clearly suggests the usefulness of unlabeled data and 
the consistent benefits of adopting semi-supervised learning for TAD -- a largely ignored research topic.
{\bf(2)} Combining existing TAD methods (\eg, BMN \cite{lin2019bmn}) with previous SSL methods 
(\eg, Mean Teacher \cite{tarvainen2017mean} and FixMatch \cite{sohn2020fixmatch}) is indeed effective in improving model generalization.
Similar observations can be made on SSP \cite{ji2019learning} and SS-TAD \cite{wang2021self}, our two main competitors.
Despite such performance improvements, 
these methods still suffer from the proposal error propagation problem that would limit their ability in
exploiting unlabeled videos.
{\bf(3)} By solving this problem with a new parallel design,
our {\shortmodelname} achieves the new state of the art on both datasets.
It is worth pointing out the larger margin achieved by {\shortmodelname}
over all the competitors at the lower supervision case (10\% label). For instance, with 10\% label, {\shortmodelname} surpasses the second best SSTAP by 3.9\%/4.3\% on ActivityNet/THUMOS.
A plausible reason is that the proposal error propagation will become more severe when the labeled set is smaller, causing more harm to existing proposal based TAD methods.
This validates the overall efficacy and capability of our model formulation
in leveraging unlabeled video data for SS-TAD.
\textcolor{black}{\textbf{(4)} From Table \ref{tab:SSL_2} we observe similar findings as that of BMN \cite{lin2019bmn} using TSN features (Table \ref{Tab:semi}). Our {\shortmodelname} outperforms the existing baselines and SS-TAD models by almost similar margin, confirming that the superiority of our method is feature agnostic.}


\subsection{Further Analysis}

\noindent{\bf Localization error propagation analysis }
To examine the effect of localization error propagation
with previous TAD models,
we design a proof-of-concept experiment 
by measuring the performance drop between ground-truth proposals
and pseudo proposals.
Due to the lack of access to the now obsolete training environment for UNet \cite{wang2017untrimmednets},
we adopt a MLP classifier with BMN \cite{lin2019bmn} as the baseline TAD model.
For our {\shortmodelname}, we contrast ground-truth and pseudo masks.
This experiment is tested on ActivityNet with 10\% supervision.
Table \ref{tab:err} shows that the proposal based
TAD baseline suffers almost double performance degradation from localization (\ie, proposal) error due to its sequential localization and classification design.
This verifies the advantage of {\shortmodelname}'s parallel design.
\begin{table}[h]
    \centering
    \vspace{-0.3in}
    \setlength{\tabcolsep}{0.3cm}
    \caption{Analysis of localization error propagation on ActivityNet with 10\% supervision. 
GT: Ground-Truth.}
           \begin{tabular}{@{}cll@{}}

\toprule
\multirow{2}{*}{Metric}                             & \multicolumn{2}{c}{mAP}       \\ \cmidrule(l){2-3} 
                                                     & 0.5           & Avg           \\ \midrule
\multicolumn{3}{c}{BMN \cite{lin2019bmn} + MLP}                                                          \\ \midrule
\multicolumn{1}{c|}{GT proposals}        & 55.7          & 45.3          \\ 
\multicolumn{1}{c|}{Pseudo proposals} & 32.4 ($\downarrow$ 21.3 \%)         & 23.6 ($\downarrow$ 27.7 \%)          \\\midrule
\multicolumn{3}{c}{\bf {\shortmodelname}}                                                          \\ \midrule
\multicolumn{1}{c|}{GT masks}
& 59.2          & 47.0          \\ 
\multicolumn{1}{c|}{{Pseudo masks}}           & {49.9} ($\downarrow$ \bf 9.3 \%) & {32.1} ($\downarrow$ \bf 14.9 \%) \\

\bottomrule
\end{tabular}
\label{tab:err}
\vspace{-0.1in}
\end{table}

\begin{figure}
    \centering
   \includegraphics[scale = 0.3]{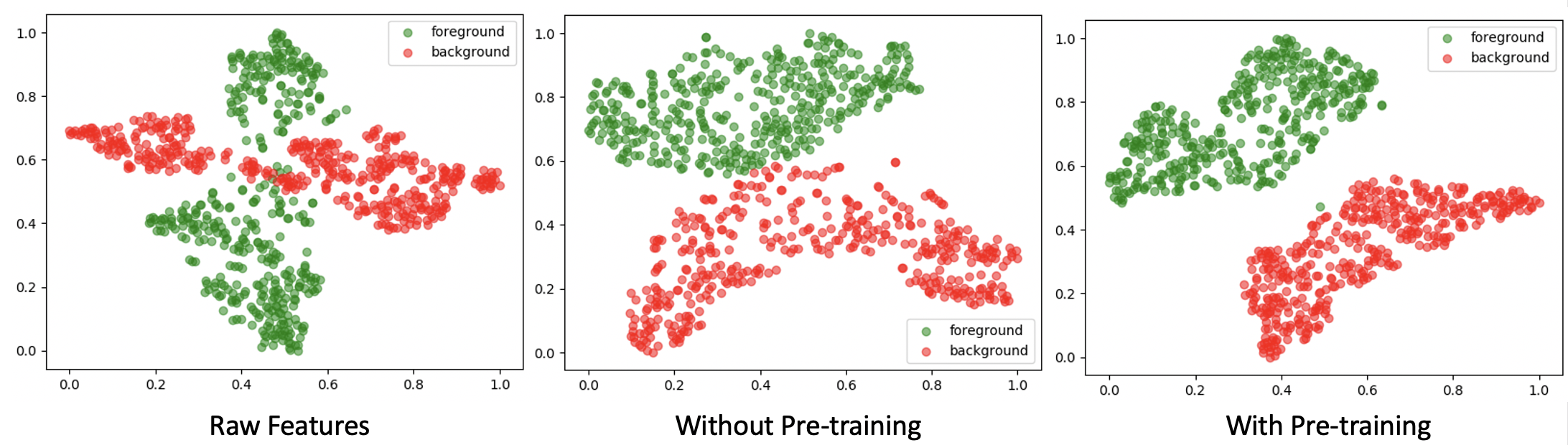}
    \caption{Effect of our model pre-training on a random ActivityNet video.}
    \label{fig:pretrain}
    \vspace{-0.5in}
\end{figure}

\noindent{\bf Effectiveness of model pre-training}
\label{sec:train}
We propose a two-staged training method characterized by 
model pre-training based on a novel pretext task (Sec. \ref{sec:train}).
We now examine how pre-training helps for model accuracy.
Here, we take the 10\% label supervision case on ActivityNet, with a comparison to
the {\rm random initialization} baseline.
Concretely, we examine three loss terms:
the mask prediction loss ($L_m$),
the feature reconstruction loss ($L_{rec}$),
and the temporal position loss ($L_{tp}$).
The results in Table \ref{Tab:unsuppre} reveal that:
{\bf (1)} Compared to random initialization, our pre-training boosts the overall mAP by $1.6\%$.
{\color{black}This gain is also supported by the more distinct separation 
of foreground and background features as
shown in Fig.~\ref{fig:pretrain}}
%
This validates the importance of TAD model initialization and the efficacy of our pretext task.
{\bf (2)} All the loss terms are effective either in isolation or in the cohorts, suggesting they are complementary with good compatibility.
\textcolor{black}{Additionally, we evaluate the generic impact of our pre-training on BMN \cite{lin2019bmn}. We observe from Table \ref{Tab:ssl_bmn} that it again gives a good increase of $0.7\%$ in Avg mAP,
justifying its generic usefulness.}

\begin{table}[]
    \centering
   \caption{Analysis of {\shortmodelname} model pre-training on ActivityNet with 10\% label supervision.
$L_m$: The mask prediction loss;
$L_{rec}$: The feature reconstruction loss;
$L_{tp}$: The temporal position loss.}
        \setlength{\tabcolsep}{0.5cm}
      \begin{tabular}{@{}ccccc@{}}
\toprule
\multicolumn{3}{c|}{Pre-training loss}                            & \multicolumn{2}{c}{mAP}       \\ \midrule
$L_{m}$      & $L_{rec}$   & \multicolumn{1}{c|}{$L_{tp}$}      & 0.5           & Avg           \\ \midrule
\multicolumn{5}{c}{W/O our pre-training (\ie, random initialization)}                                                                    \\ \midrule
\xmark          & \xmark          & \multicolumn{1}{c|}{\xmark}          & 46.2          & 30.5          \\ \midrule
\multicolumn{5}{c}{W/o our pre-training}                                                                       \\ \midrule
\cmark          & \xmark          & \multicolumn{1}{c|}{\xmark}          & 47.6          & 31.4          \\
\cmark          & \cmark          & \multicolumn{1}{c|}{\xmark}          & 48.5          & 31.7          \\
\cmark          & \xmark          & \multicolumn{1}{c|}{\cmark}          & 47.9          & 31.5          \\ \midrule
$\textbf{\cmark}$ & $\textbf{\cmark}$ & \multicolumn{1}{c|}{$\textbf{\cmark}$} & $\textbf{49.9}$ & $\textbf{32.1}$ \\ \bottomrule
\end{tabular}
\label{Tab:unsuppre}
\end{table}

\noindent{\bf Effectiveness of using unlabeled data}
\label{sec:ssl}
\textcolor{black}{We evaluate the impact of using unlabeled data imposed by the pre-training and the loss term $L_{c}$ (in Eq. \eqref{eq:final_clf}).
Table \ref{Tab:no_ssl} shows that without these components, the model will degrade in performance particularly in case of less labels, as expected. 
\textcolor{black}{In particular, both SSL components (pretraining and $L_{c}$) are clearly effective in model performance. This verifies their designs. Besides, even without the SSL modules our model is better than SSTAP \cite{wang2021self} in 10\% label case and comparable in 60\% label case, 
further indicating the advantage of our model design.
}
}

\begin{table}[h]
	\begin{minipage}{0.5\linewidth}
		\centering
		\caption{\textbf{SS-TAD results \\ on THUMOS14 test-set} using I3D features; All method except \textbf{{\shortmodelname}} uses UNet \cite{wang2017untrimmednets}; $^{*}$:Using only labeled set.}
         \label{tab:SSL_2}
         
         \resizebox{\textwidth}{!}{
         \setlength{\tabcolsep}{0.1cm}
\begin{tabular}{@{}ccccc@{}}
\toprule
\multirow{2}{*}{Labels}                    & \multirow{2}{*}{Method}              & \multicolumn{3}{c}{mAP}                 \\ \cmidrule(l){3-5} 
                                           &                                      & 0.3         & 0.5         & 0.7         \\ \midrule
\multicolumn{1}{c|}{\multirow{4}{*}{60\%}} & \multicolumn{1}{c|}{$GTAD^{*}$\cite{xu2020boundary}}            & 50.9        & 35.4        & 16.5        \\ \cmidrule(l){2-5} 
\multicolumn{1}{c|}{}                      & \multicolumn{1}{c|}{FixMatch \cite{sohn2020fixmatch}+GTAD} & 53.4        & 38.9        & 19.1        \\
\multicolumn{1}{c|}{}                      & \multicolumn{1}{c|}{SSP \cite{ji2019learning}}      & 53.5        & 39.7        & 20.4        
\\
\multicolumn{1}{c|}{}                      & \multicolumn{1}{c|}{SSTAP\cite{wang2021self}+GTAD}      & 55.9        & 41.6        & 22.0        \\ \cmidrule(l){2-5} 
\multicolumn{1}{c|}{} & \multicolumn{1}{c|}{\textbf{{\shortmodelname} (Ours)}} & \textbf{58.7} & \textbf{42.4} & \textbf{23.1} \\ \midrule
\multicolumn{1}{c|}{\multirow{4}{*}{10\%}} & \multicolumn{1}{c|}{$GTAD^{*}$\cite{xu2020boundary}}            & 36.9        & 20.1        & 6.6         \\ \cmidrule(l){2-5} 
\multicolumn{1}{c|}{}                      & \multicolumn{1}{c|}{FixMatch \cite{sohn2020fixmatch}+GTAD} & 42.1        & 23.8        & 9.7         \\
\multicolumn{1}{c|}{}                      & \multicolumn{1}{c|}{SSP \cite{ji2019learning}}      & 43.1        & 25.5        & 9.6        \\
\multicolumn{1}{c|}{}                      & \multicolumn{1}{c|}{SSTAP \cite{wang2021self}+GTAD}      & 45.3        & 27.5        & 11.0        \\\cmidrule(l){2-5} 
\multicolumn{1}{c|}{}                      & \multicolumn{1}{c|}{\textbf{{\shortmodelname} (Ours)} }                & \textbf{49.1} & \textbf{31.7} & \textbf{12.6} \\ \bottomrule
\end{tabular}
}
	\end{minipage}\hfill
	\begin{minipage}{0.45\linewidth}
	 \caption{Performance of our SPOT model w/ and  w/o unlabeled data on ActivityNet.}
\setlength{\tabcolsep}{0.2cm}
\begin{tabular}{@{}c|cc|cc@{}}
\toprule

\multirow{2}{*}{Labels} & \multicolumn{2}{c|}{SSL Modules}                         & \multicolumn{2}{c}{mAP}                            \\ \cmidrule(l){2-5} 
                        & \multicolumn{1}{c|}{Pre-train} & $L_{c}$ & \multicolumn{1}{c|}{0.5}           & Avg           \\ \midrule
\multirow{4}{*}{10\%}  & \multicolumn{1}{c|}{\cmark}            & \cmark                    & \multicolumn{1}{c|}{\textbf{49.9}} & \textbf{32.1} \\ \cmidrule(l){2-5} 
                        & \multicolumn{1}{c|}{\cmark}            & \xmark                   & \multicolumn{1}{c|}{45.3}          & 29.8          \\
                        & \multicolumn{1}{c|}{\xmark}            & \cmark                    & \multicolumn{1}{c|}{46.2}          & 30.5          \\
                        & \multicolumn{1}{c|}{\xmark}            & \xmark                    & \multicolumn{1}{c|}{44.5}          & 28.3          \\ \midrule
\multirow{4}{*}{60\%}  & \multicolumn{1}{c|}{\cmark}            & \cmark                    & \multicolumn{1}{c|}{\textbf{52.8}} & \textbf{35.2} \\ \cmidrule(l){2-5} 
                        & \multicolumn{1}{c|}{\cmark}            & \xmark                    & \multicolumn{1}{c|}{51.7}          & 34.4          \\
                        & \multicolumn{1}{c|}{\xmark}            & \cmark                    & \multicolumn{1}{c|}{52.1}          & 34.9          \\
                        & \multicolumn{1}{c|}{\xmark}            & \xmark                    & \multicolumn{1}{c|}{51.2}          & 34.0          \\ \bottomrule
\end{tabular}
\label{Tab:no_ssl}
	\end{minipage}
	  \vspace{-0.2in}
\end{table}

\noindent{\bf Usefulness of boundary refinement }
In Sec.~\ref{sec:tal} we introduced a TAD refinement component based on inter-stream interaction with the aim to mitigate the boundary ambiguity problem.
Here we examine its effectiveness in performance boost
on ActivityNet with 10\% labeled data. 
Besides the entire benefit,
we also test the respective effects of the background and foreground terms, \textcolor{black}{corresponding to the first and second loss terms}
of Eq. \eqref{eq:ref}.
Table \ref{Tab:refine} show that:
{\bf (1)} Our inter-stream interaction is effective for TAD refinement in semi-supervised learning with 1.5\% gain in mAP.
This verifies our consideration on the importance of boundary inference
and our model design.
{\bf (2)} Both foreground and background terms are useful in isolation, and importantly present strong synergy as their combined effect (1.5\% mAP gain) is much greater than the sum of their individual gains (0.7\%=0.5\%+0.2\%).
This is not surprising as the boundary reasoning requires to model both foreground and background 
well simultaneously.
\begin{table}[h]
    \centering
    \vspace{-0.3in}
    \setlength{\tabcolsep}{0.4cm}
    \caption{Analysis of inter-stream interaction for TAD refinement on ActivityNet with 10\% label supervision.}
            \begin{tabular}{@{}cc|cc@{}}
\toprule
\multicolumn{2}{c|}{Refinement loss (Eq. \eqref{eq:ref})} & \multicolumn{2}{c}{mAP} \\ \midrule
Foreground Term    &  Background Term  & 0.5        & Avg        \\ \midrule
\xmark          & \xmark           & 46.8       & 30.7       \\ \midrule
\cmark          & \cmark           & \bf 49.9       & \bf 32.1       \\ \midrule
\xmark          & \cmark           & 47.2       & 31.1       \\
\cmark          & \xmark           & 46.9       & 30.8       \\
\bottomrule
\end{tabular}
\label{Tab:refine}
\vspace{-0.1in}
\end{table}

\noindent{\bf Contributions of different losses}
We ablate the effect of {\shortmodelname}'s training loss functions
for classification and mask (Sec. \ref{sec:train_s2}).
We test the benefit of 
the classification loss ($L_c$),
the weighted binary cross-entropy mask loss ($L_{mce}$),
the dice mask loss ($L_{dice}$),
the refinement loss ($L_{ref}$), and
the feature reconstruction loss ($L_{rec}$).
When testing $L_c$, we replace it with the standard cross-entropy loss.
Table \ref{tab:loss} shows that 
each loss term is beneficial for improving model accuracy,
verifying their individual and collective efficacy.

\noindent{\bf Comparisons to fully supervised TAD methods }
Besides SS-TAD, our {\shortmodelname} can be also applied for fully supervised TAD with the pseudo-labels replaced by ground-truth labels while keeping the remaining the same.
This test is conducted on ActivityNet.
Table \ref{tab:sup} shows that when trained with fully-labeled data, our {\shortmodelname} can also outperform state-of-the-art TAD methods
in the overall result albeit by a smaller margin as compared to the semi-supervised case.
This is as expected because in fully supervised setting there would be less proposal error and
hence less harm by its propagation.


\begin{table}[!htb]
     \vspace{-0.1in}
     \setlength{\tabcolsep}{0.5cm}
    \begin{minipage}{.45\linewidth}
    \centering
      \caption{
      Ablating {\shortmodelname} objective loss terms on ActivityNet with 10\% label.
$L_c$: classification loss;
$L_{mce}$: cross-entropy mask loss;
$L_{dice}$: dice mask loss;
$L_{ref}$: refinement loss;
$L_{rec}$: feature reconstruction loss.
      }
      \resizebox{1.0\textwidth}{!}{
      \centering
              \begin{tabular}{@{}l|cc@{}}
\toprule
{\multirow{2}{*}{\textbf{Loss}}} & \multicolumn{2}{c}{\textbf{mAP}}                \\ \cmidrule(l){2-3} 
\multicolumn{1}{c|}{}                               & {\textbf{0.5}} & \textbf{Avg} \\ \midrule
{\bf All}                     & {\bf 49.9}           & \bf 32.1           \\ \midrule
{W/O  $L_c$}           & {45.3}           & 29.8           \\
{W/O $L_{mce}$}                   & {47.8}           & 31.5           \\
{W/O $L_{dice}$}                                & {47.0}                                & 31.2           \\
{W/O $L_{rec}$}                                & {45.9}                                & 30.2           \\ 
{W/O $L_{ref}$}                                & {46.8}                                & 30.7           \\ \bottomrule
\end{tabular}
\label{tab:loss}
        }
    \end{minipage}%
    \hfill
    \begin{minipage}{.5\linewidth}
      \centering
        \caption{\textbf{Fully supervised TAD results on the validation set of
ActivityNet.} All the compared methods use TSN features used in \cite{wang2021self}}
\label{tab:sup}
           \setlength{\tabcolsep}{0.06cm}
\resizebox{\textwidth}{!}{
\begin{tabular}{@{}c|cccc@{}}
\toprule
\multirow{2}{*}{Method} & \multicolumn{4}{c}{mAP}                                                           \\ \cmidrule(l){2-5} 
                        & 0.5           & 0.75          & \multicolumn{1}{c|}{0.95}         & Avg           \\ \midrule
BSN \cite{lin2018bsn}                    & 46.4          & 29.9          & \multicolumn{1}{c|}{8.0}          & 30.0          \\
GTAD \cite{xu2020g}                    & 50.3          & 34.6          & \multicolumn{1}{c|}{9.0}          & 34.0          \\
BC-GNN \cite{bai2020boundary}                  & 50.6          & 34.8          & \multicolumn{1}{c|}{\bf 9.4}          & 34.3          \\
BMN \cite{lin2019bmn}              & 50.0          & 34.8          & \multicolumn{1}{c|}{8.3}          & 33.8          \\
 
BSN++ \cite{su2020bsn++}                 & 51.2         & 35.7           & \multicolumn{1}{c|}{8.3}          & 34.8 \\
TCANet \cite{Qing_2021_CVPR}                 & 52.2          & 36.7          & \multicolumn{1}{c|}{6.8}          & 35.5 \\   
GTAD+CSA\cite{sridhar2021class}                 & 51.8          & \textbf{36.8}          & \multicolumn{1}{c|}{8.7}          & 35.6 
\\
\midrule

\textbf{{\shortmodelname}}           & \textbf{53.9} & 35.9 & \multicolumn{1}{c|}{\textbf{9.4}} & \textbf{35.8} \\ \bottomrule
\end{tabular}
} 
    \end{minipage} 
      \vspace{-0.1in}
\end{table}

\noindent{\bf Training and inference complexity }
We compare
{\shortmodelname} with a representative TAD method BMN \cite{lin2019bmn} with our pre-training and a recent SS-TAD method SSTAP \cite{xu2020g}. 
All the methods are tested under the same training setting and the same machine with one Nvidia 2080 Ti GPU.
We measure the entire training time (including pre-training)
and average inference time per video in testing.
{We use the two-stream video features used in \cite{wang2021self}}
It can be seen in Table \ref{tab:time} that despite with pre-training and
fine-tuning our {\shortmodelname} is still drastically faster, \eg, $31/35\times$ for
training and $2.3\times$ for testing in comparison to SSTAP/BMN, respectively.
This validates our motivation of designing a proposal-free SS-TAD model
in terms of computational efficiency.

\begin{table}[t]
\small
    \setlength{\tabcolsep}{0.1cm}
    \begin{minipage}{.45\linewidth}
      \caption{
      Comparison of training and inference time on a single gpu.
}

      \centering
\begin{tabular}{@{}c|cc|c@{}}
\toprule
\multirow{2}{*}{Method} & \multicolumn{2}{c|}{Training (in hrs)} & \multirow{2}{*}{\begin{tabular}[c]{@{}c@{}}Inference\\ (in secs)\end{tabular}} \\ \cmidrule(lr){2-3}
            & Pre-train & Fine-tune &      \\ \midrule
BMN \cite{lin2019bmn}     & 4.0          & 6.2        & 0.21 \\
SSTAP \cite{wang2021self}        & -          & 9.4        & 0.21 \\

\textbf{{\shortmodelname}} & \textbf{0.10}         & \textbf{0.21}       & \textbf{0.09} \\ \bottomrule
\end{tabular}
\label{tab:time}
    \end{minipage}%
    \hfill
    \begin{minipage}{.4\linewidth}
      \centering
      \setlength{\tabcolsep}{0.2cm}
        \caption{Effect of pre-training on ActivityNet with 10\% labels.}
       \begin{tabular}{@{}c|cc@{}}
\toprule
\multirow{2}{*}{Method} & \multicolumn{2}{c}{mAP}          \\ \cmidrule(l){2-3} 
                        & \multicolumn{1}{c|}{0.5}  & Avg  \\ \midrule
BMN + UNet \cite{wang2017untrimmednets}                   & \multicolumn{1}{c|}{35.4} & 25.8 \\
$BMN^{\dagger}$ + UNet \cite{wang2017untrimmednets}                   & \multicolumn{1}{c|}{\textbf{36.2}} & \textbf{26.3} \\ \bottomrule
\end{tabular}
\label{Tab:ssl_bmn}
\end{minipage}

\end{table}

\noindent\textbf{Role of Positional Encoding }
\textcolor{black}{We evaluate the effect of position encoding on ActivityNet. 
As shown in Table \ref{tab:posenc}, it is interesting to see that position encoding is not necessary and even harmful to the performance. This indicates that with our current formulation, the snippet level temporal information does not bring extra useful information.}

\noindent\textbf{Effect of Pre-training Loss in Finetuning}
\textcolor{black}{
Recall that we do not use the pre-training loss $L_{tp}$ (for temporal ordering pretext task) during the finetuning stage as shown in Sec 3.3. 
Table \ref{tab:preloss} shows a slight drop of $0.2\%$ in avg mAP from this pre-text loss which may be due to the incompatibility with the classification loss $L_{c}$ of TAD. This is not rare in pretraining-finetuning pipeline
with the pretext loss dropped during finetuning.}

\begin{table}[!htbp]
\small
    \renewcommand{\tabcolsep}{15pt}
    \begin{minipage}{.45\linewidth}
      \caption{Positional Encoding on \\ ActivityNet W/ 10\% labels.}
      \label{tab:posenc}
      \centering
\begin{tabular}{@{}l|cc@{}}
\toprule
\multirow{2}{*}{\begin{tabular}[c]{@{}c@{}}\# Position\\ Encoding\end{tabular}} & \multicolumn{2}{c}{mAP} \\ \cmidrule(l){2-3} 
              & \multicolumn{1}{c|}{0.5}  & Avg  \\ \midrule
No Encoding   & \multicolumn{1}{c|}{\bf 49.9} &\bf 32.1 \\
Learnable       & \multicolumn{1}{c|}{46.7} & 29.4 \\
Non-Learnable & \multicolumn{1}{c|}{39.8} & 24.2 \\ \bottomrule
\end{tabular}
    \end{minipage}%
    \hfill
    \begin{minipage}{.5\linewidth}
      \centering
      \renewcommand{\tabcolsep}{30pt}
        \caption{Ablation of pretraining loss \\ during finetuning W/ 10\% labels  \\ on ActivityNet.}
        \label{tab:preloss}
\begin{tabular}{@{}c|c@{}}
\toprule
Loss               & Avg  \\ \midrule
W/O  $L_{tp}$      & \bf 32.1 \\ \midrule
W/ $L_{tp}$    &  31.9 \\
\bottomrule
\end{tabular}
\end{minipage} 
\vspace{-0.4cm}
\end{table}

\begin{figure}
    \centering
    \includegraphics[scale=0.3]{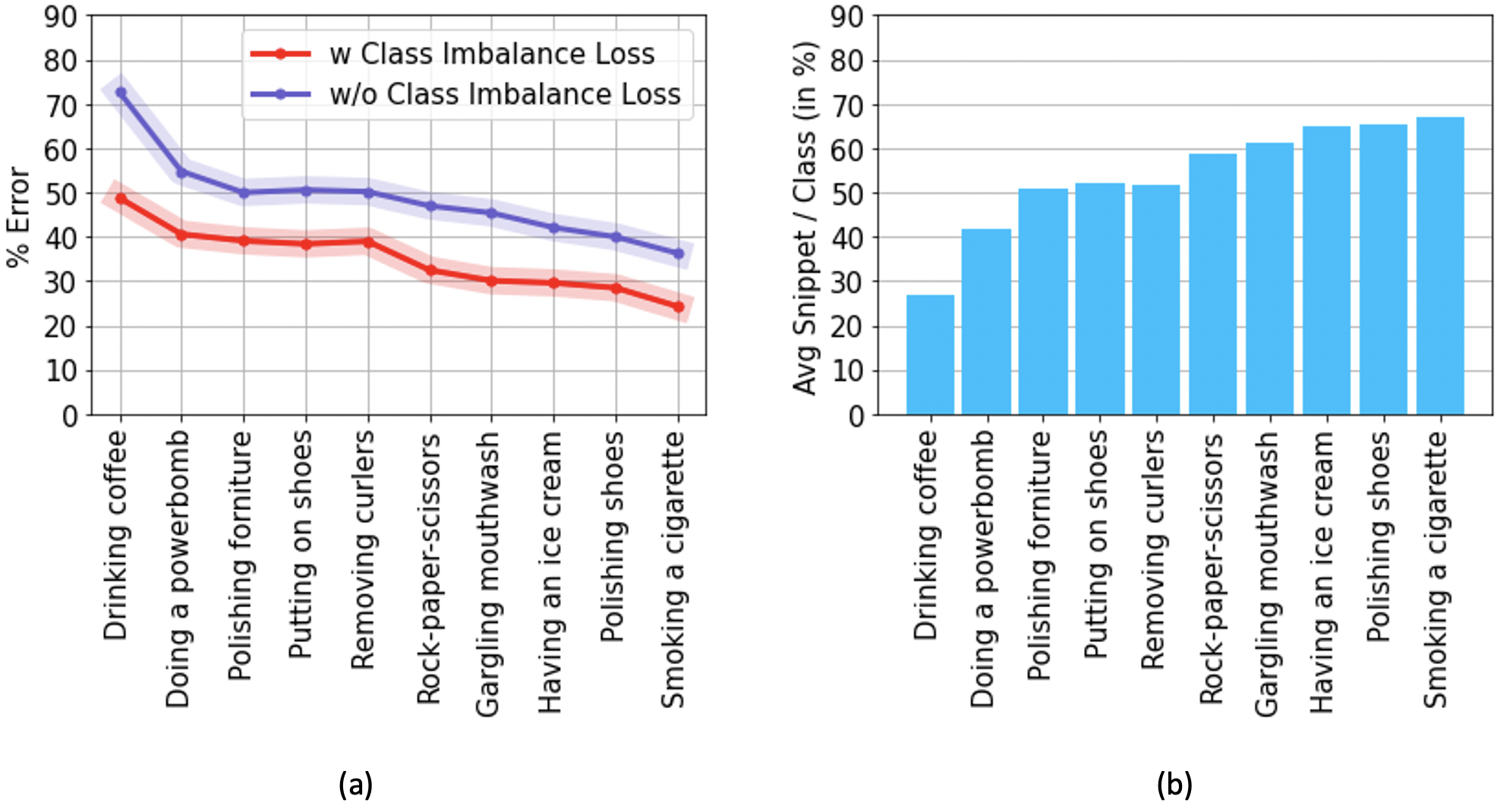}
    \caption{\textbf{Effect of tackling class imbalance} on  top-10 tail classes from ActivityNet. (a) shows the effect of dealing with the class imbalance. (b) shows the foreground coverage per class.}
    \label{fig:imbalance}
\end{figure}

\noindent\textbf{Handling Class-Imbalance Challenge }
\textcolor{black}{In section 3.3 of main paper, we introduce a new class-balanced loss to handle the class-imbalance problem in SS-TAD. We evaluate on the videos corresponding to top-10 tail classes on ActivityNet in terms of error rate. 
The imbalance problem arises mainly due to the snippet coverage of a particular class (Fig. ~\ref{fig:imbalance}(b)),
and we see a high correlation  with the error rate in Fig. ~\ref{fig:imbalance}(a).
Importantly, using our class-balanced loss,  the error rate of the heavily imbalanced \textit{``Drinking Coffee''} class can be reduced by $\sim 20\%$.}

\section{Qualitative Analysis}
\begin{figure*}[!htbp]
\begin{center}
  \includegraphics[scale=0.24]{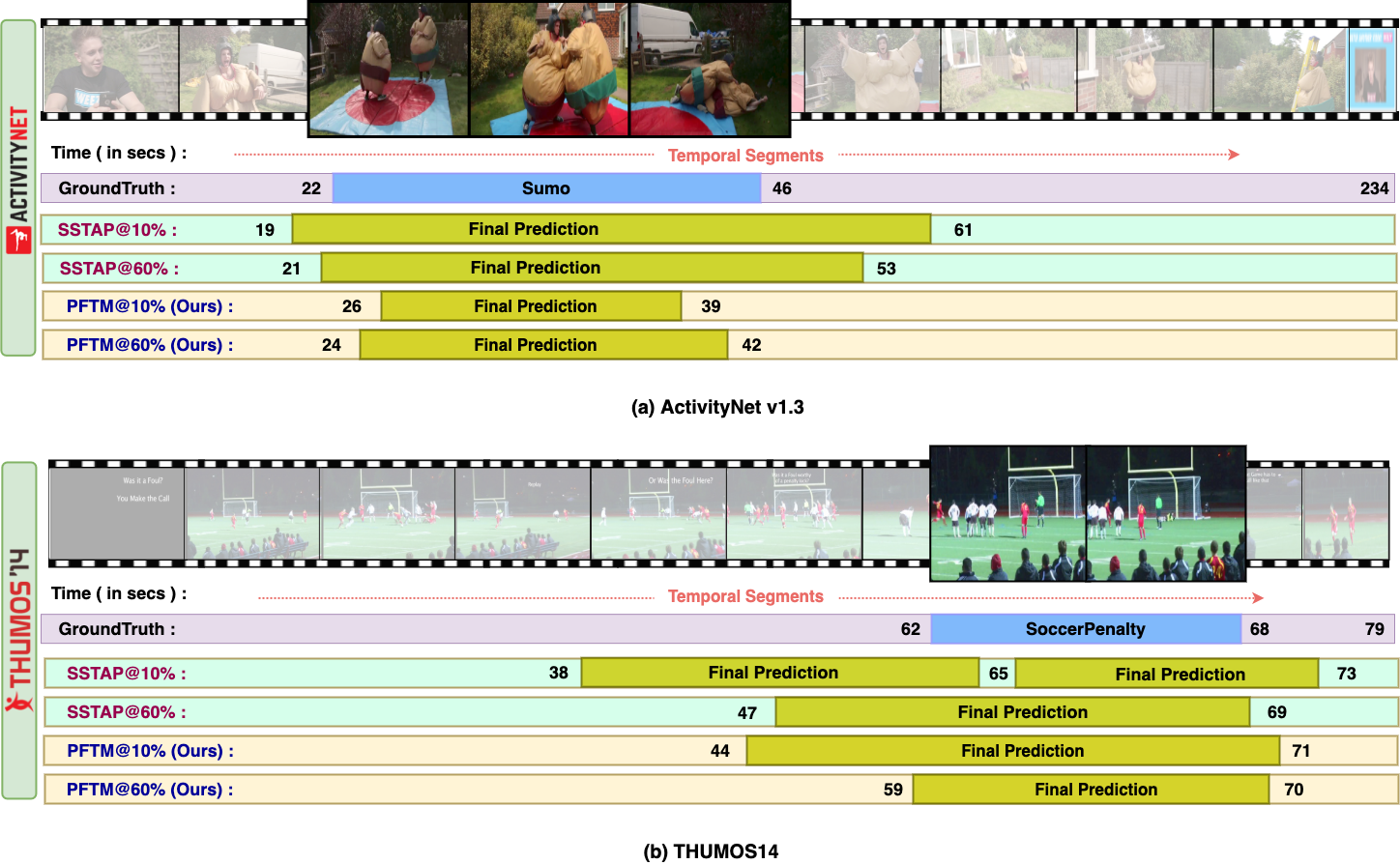}
\end{center}
\caption{\textbf{Qualitative SS-TAD result comparison} on two untrimmed action videos from (a) ActivityNet and (b) Thumos. 
}
\label{fig:example1}
\end{figure*}

In this section, to make more visual examination we provide additional qualitative results by SSTAP \cite{wang2021self} and our {\shortmodelname} model for $10\%$ and $60\%$ available data variant on both ActivityNet and THUMOS dataset. From the illustration in Fig. \ref{fig:example1}, we have a similar observation that compared to SSTAP, our proposed {\shortmodelname} method can localize the target action instances more accurately at both $10\%$ and $60\%$ labeled data and as we increase the labels the detection quality improves. This finding is consistent across the two datasets.

\section{Limitation}
\textcolor{black}{ In situations where duration of background instances in between two foreground instances are small, {\shortmodelname} may wrongly consider the background in between as foreground. Since snippet is the smallest prediction unit, any foreground/background segments with a duration close to the snippet length will challenge any TAD methods that use snippets as input.  Improving the sensitivity of our model on detecting short-duration action instances will be part of future work.}

\section{Conclusions}
In this work, we have proposed a novel {\em Proposal-Free Temporal Mask} ({\shortmodelname}) learning model
for the under-studied yet practically useful semi-supervised temporal action detection (SS-TAD).
It is characterized by
a {\em parallel}
localization (mask generation) and classification architecture
designed to solve the localization error propagation problem
with conventional TAD models.
For superior optimization, we further introduced a novel pretext task for pre-training and a boundary refinement algorithm.
Extensive experiments on ActivityNet and THUMOS demonstrate that our {\shortmodelname} yields state-of-the-art performance under both semi-supervised and supervised learning settings.

%
%
\bibliographystyle{splncs04}
\bibliography{egbib}
\end{document}